\documentclass[twoside]{article}
\usepackage[accepted]{aistats2019}
\PassOptionsToPackage{usenames,dvipsnames,dvinames}{xcolor}
\usepackage[usenames,dvipsnames,dvinames]{xcolor}
\usepackage{multirow}
\usepackage{microtype}
\usepackage{enumitem}
\usepackage{amsmath,amsthm}
\usepackage[mathscr]{eucal}
\usepackage{amssymb}
\usepackage{footnote}
\usepackage{graphicx}
\usepackage{caption}
\usepackage{subfloat}
\usepackage[vlined,ruled]{algorithm2e}
\usepackage{algorithmic}
\usepackage{wrapfig}
\usepackage{float}
\usepackage{caption}
\usepackage{titlesec}
\titlespacing\subsection{0pt}{4pt plus 2pt minus 2pt}{4pt plus 2pt minus 2pt}

\newcommand{\lovasz}{Lov\'asz}

\DeclareMathOperator*{\argmax}{argmax}

\providecommand{\doarxiv}{false}
\usepackage{ifthen}
\newboolean{isarxiv}
\setboolean{isarxiv}{\doarxiv} 
\ifthenelse{\boolean{isarxiv}}{%
\newcommand{\arxiv}[1]{#1}
\newcommand{\notarxiv}[1]{}
}{
\newcommand{\arxiv}[1]{}
\newcommand{\notarxiv}[1]{#1}
}
\newcommand{\arxivalt}[2]{\ifthenelse{\boolean{isarxiv}}{#1}{#2}}
\newcommand{\arxivaltr}[2]{\ifthenelse{\boolean{isarxiv}}{#2}{#1}}

\usepackage{ifthen}
\newboolean{isdraft}
\setboolean{isdraft}{false} 
\ifthenelse{\boolean{isdraft}}{%
\newcommand{\myaddcomment}[3]{{\color{#1}{\ensuremath{\langle\!\!\langle}{\bf {#2} :} {#3}\ensuremath{\rangle\!\!\rangle}}}}
\newcommand{\rishabh}[1]{\myaddcomment{orange}{Rishabh}{#1}}
\newcommand{\JTR}[1]{\myaddcomment{orange}{Jeff\ensuremath{\rightarrow}Rishabh}{#1}}
\newcommand{\jeff}[1]{\myaddcomment{blue}{Jeff}{#1}}
\newcommand{\RTJ}[1]{\myaddcomment{blue}{Rishabh\ensuremath{\rightarrow}Jeff}{#1}}
}{
\newcommand{\rishabh}[1]{}
\newcommand{\JTR}[1]{}
\newcommand{\jeff}[1]{}
\newcommand{\RTJ}[1]{}
\newcommand{\toboth}[1]{}
}

\usepackage[bookmarks, colorlinks=true, citecolor=Violet,linkcolor=Mahogany,urlcolor=blue]{hyperref}

\begin{document}

\twocolumn[
\aistatstitle{A Memoization Framework for Scaling Submodular Optimization to Large Scale Problems}

\aistatsauthor{ Rishabh Iyer \And Jeff Bilmes }

\aistatsaddress{ Microsoft Corporation \And  University of Washington, Seattle}
]

\begin{abstract}
We are motivated by large scale submodular optimization problems, where standard algorithms that treat the submodular functions in the \emph{value oracle model} do not scale. In this paper, we present a model called the \emph{precomputational complexity model}, along with a unifying memoization based framework, which looks at the specific form of the given submodular function. A key ingredient in this framework is the notion of a \emph{precomputed statistic}, which is maintained in the course of the algorithms. We show that we can easily integrate this idea into a large class of submodular optimization problems including constrained and unconstrained submodular maximization, minimization, difference of submodular optimization, optimization with submodular constraints and several other related optimization problems. Moreover, memoization can be integrated in both discrete and continuous relaxation flavors of algorithms for these problems. We demonstrate this idea for several commonly occurring submodular functions, and show how the precomputational model provides significant speedups compared to the value oracle model. Finally, we empirically demonstrate this for large scale machine learning problems of data subset selection and summarization.\looseness-1
\end{abstract}

\section{Introduction} \label{sec:introduction}
Submodular functions provide a rich class of expressible models for a variety of machine learning problems. Submodular functions occur naturally in two flavors. In minimization problems, they model notions of cooperation, attractive potentials and economies of scale, while in maximization problems, they model aspects of coverage, diversity and information. As a result, they have repeatedly appeared in several real world problems including document summarization~\cite{linacl}, image summarization~\cite{tschiatschek2014learning}, data subset
selection and active learning~\cite{lin2012submodularity,wei2015submodularity}, image segmentation and denoising~\cite{frbach1, curvaturemin,jegelka2011-nonsubmod-vision} and many others. A set function
$f: 2^V \to \mathbb R$ over a finite set $V = \{1, 2, \ldots, n\}$ is
\emph{submodular} if for all subsets $S, T \subseteq V$, it holds that
$f(S) + f(T) \geq f(S \cup T) + f(S \cap T)$. Given a set $S \subseteq
V$, we define the \emph{gain} of an element $j \notin S$ in the
context $S$ as $f(j | S) \triangleq f(S \cup j) - f(S)$. A more intuitive characterization is the diminishing returns characterization: A function $f$ is
submodular if it satisfies \emph{diminishing marginal returns},
namely $f(j | S) \geq f(j | T)$ for all $S \subseteq T, j \notin T$,
and is \emph{monotone} if $f(j | S) \geq 0$ for all $j \notin S, S
\subseteq V$.

While submodular functions naturally occur in a number of real world applications, they also admit nice theoretical characterizations and algorithms. In particular, many simple iterative procedures like greedy~\cite{nemhauser1978}, local search~\cite{janvondrak} and majorization-minimization~\cite{rkiyersemiframework2013} yield theoretical guarantees for these problems. These algorithms are very efficient, scalable and easy to implement, and hence are being used more often in several large scale machine learning problems. The scale of machine learning problems are often massive, with dataset sizes of several hundreds of millions of examples. This has led to significant research in providing distributive, streaming, and multi-stage procedures for scaling these problems~\cite{mirzasoleiman13distributed,wei2014fast, badanidiyuru2014fast}. 

Many existing submodular optimization algorithms, treat the submodular function as a black box -- a model, called the \emph{value oracle model}. While this yields a useful way of quantifying the complexity of these algorithms, it does not provide an efficient way of implementing real world instances of submodular functions, which all appear in succinct representation, and can be can be stored in time and space, polynomial in the size of the ground set. A few recent papers have attempted to go beyond the value oracle model. For example, \cite{hassidim2016submodular, horel2016maximization, singla2016noisy} study submodular optimization with noisy oracles. Similarly~\cite{karimi2017stochastic, asadpour2008stochastic, mokhtari2017conditional, hassani2017gradient} study scenarios where we don't have the closed form expression of the submodular functions, but have a stochastic approximation (available for example through simulations). Similarly~\cite{stan2017probabilistic, balkanski2016learning} study the problem of maximizing submodular functions drawn from specific probability distributions. 

In this paper, we take an orthogonal direction by introducing a new complexity model called the \emph{precomputational complexity model}, along with a unifying memoization framework. We define the notion of a \emph{precomputed statistic}, which is specific to a submodular function, and can be integrated easily into a large class of existing submodular optimization problems and algorithms.
We then show how to compute these statistics for several real world submodular functions occurring in applications, and theoretically demonstrate how the precomputational model reveals improved complexity results in comparison to the value oracle model. Finally we consider a real world application of speech corpus summarization and data subset selection, and show the scalability of our framework.\looseness-1

The idea of using precomputational statistics for speeding up submodular optimization in itself is not new. For example, \cite{krause2008efficient, krause2010sfo} introduce the idea of incremental evaluation for the greedy algorithms for specific submodular functions such as entropy and mutual information. Similar ideas have also been discussed for Determinantal Point Processes~\cite{kulesza2012determinantal}. In this work, we show that we can provide a unified framework via the construct of a \emph{precomputed statistics} and show how this extends to both, a large class of submodular functions as well as a large class of optimization algorithms.
\section{Basic Ideas and Background}
\label{basicideas}
We first introduce several key concepts we shall use in this paper. We start out by defining some fundamental characteristics of submodular functions.

\textbf{The Submodular Polyhedron and \lovasz{} extension: }
For a submodular function $f$, the submodular polyhedron $\mathcal P_f$ and the corresponding base polytope $\mathcal B_f$ are respectively defined as:  \looseness-1
\begin{align} 
\mathcal P_f = \{ x : x(S) \leq f(S), \forall S \subseteq V \} 
\;\;\; \\
\mathcal B_f = \mathcal P_f \cap \{ x : x(V) = f(V) \}.
\end{align}
For a vector $x \in \mathbb{R}^V$ and a set $X \subseteq V$, we write $x(X)
= \sum_{j \in X} x(j)$. Though $\mathcal P_f$ is defined via $2^n$ inequalities, its extreme point can be easily characterized~\cite{fujishige2005submodular}. Given any permutation $\sigma$ of the ground set $\{1, 2, \cdots, n\}$, and an associated chain $\emptyset = S^{\sigma}_0 \subseteq S^{\sigma}_1 \subseteq \cdots \subseteq S^{\sigma}_n = V$ with $S^{\sigma}_i =
\{ \sigma(1), \sigma(2), \dots, \sigma(i) \}$, a vector $h^f_{\sigma}$ satisfying,
 \begin{align}\label{extremeptdef}
 h^f_{\sigma}(\sigma(i) = f(S^{\sigma}_i) - f(S^{\sigma}_{i-1}) = f(\sigma(i) | S^{\sigma}_{i-1}), \nonumber \\ \forall i = 1, \cdots, n
 \end{align}
forms an extreme point of $\mathcal P_f$. Moreover, a natural convex extension of a submodular function, called the \lovasz{} extension~\cite{lovasz1983} is closely related to the submodular polyhedron, and is defined as $\hat{f}(x) = \max_{h \in \mathcal P_f} \langle h, x \rangle$. Thanks to the properties of the polyhedron, $\hat{f}(x)$ can be efficiently computed: Denote $\sigma_x$ as an ordering induced by $x$, such that $x(\sigma_x(1)) \geq x(\sigma_x(2)) \geq \cdots x(\sigma_x(n))$. Then the \lovasz{} extension is $\hat{f}(x) = \langle h^f_{\sigma}, x \rangle$~\cite{lovasz1983}.

\textbf{Modular lower bounds (Subgradients): } Akin to convex functions, submodular functions have tight modular lower bounds. These bounds are related to the subdifferential $\partial_f(Y)$ of the submodular set function $f$ at a set $Y \subseteq V$, which is defined 
\cite{fujishige2005submodular}
as:
\begin{align*}
\partial_f(Y) = \{y \in \mathbb{R}^n: f(X) - y(X) \geq f(Y) - y(Y),\;\nonumber \\ \text{for all } X \subseteq V\}
\end{align*}
Denote a subgradient at $Y$ by $h_Y \in \partial_f(Y)$. The extreme points of
$\partial_f(Y)$ may be computed in a manner similar to those of the submodular polyhedron. Let $\sigma_Y$
be a permutation of $V$ that assigns the elements in $Y$ to the first
$|Y|$ positions ($\sigma_Y(i) \in Y$ if and only if $i \leq  |Y|$). Then, $h_Y = h^f_{\sigma_Y}$ (where $h^f_{\sigma_Y}$ is as defined in eqn.~\eqref{extremeptdef}) forms a lower bound of $f$, tight at $Y$ --- i.e.,
$h_Y(X) = \sum_{j \in X} h_Y(j) \leq f(X), \forall X
\subseteq V$ and $h_Y(Y) = f(Y)$. Notice that the extreme points of a subdifferential are a subset of the extreme points of the submodular polyhedron.\looseness-1

\textbf{Modular upper bounds (Supergradients):} We can also define superdifferentials $\partial^f(Y)$ of a submodular
function 
\cite{jegelka2011-nonsubmod-vision,rkiyersubmodBregman2012}
at
$Y$: 
\begin{align*}\label{supdiff-def}
\partial^f(Y) = \{y \in \mathbb{R}^n: f(X) - y(X) \leq f(Y) - y(Y); \nonumber \\ \text{for all } X \subseteq V\}
\end{align*}
It is possible, moreover, to provide specific supergradients~\cite{rkiyersubmodBregman2012,rkiyersemiframework2013} that define the following two modular upper bounds (when referring either one, we use $m^f_X$):
\notarxiv{\small}
\begin{align*}
m^f_{X, 1}(Y) \triangleq f(X) - \!\!\!\! \sum_{j \in X \backslash Y } f(j| X \backslash j) + \!\!\!\! \sum_{j \in Y \backslash X} f(j| \emptyset)\scalebox{1.3}{,}\;\;\; \\
m^f_{X, 2}(Y) \triangleq f(X) - \!\!\! \sum_{j \in X \backslash Y } f(j| V \backslash j) + \!\!\!\! \sum_{j \in Y \backslash X} f(j| X). \nonumber
\end{align*}
\normalsize
Then $m^f_{X, 1}(Y) \geq f(Y)$ and $m^f_{X, 2}(Y) \geq f(Y), \forall Y \subseteq V$ and $m^f_{X, 1}(X) = m^f_{X, 2}(X) = f(X)$.

\begin{table*}
\begin{center}
\small{
 \begin{tabular}{|| c | c | c |  c | c ||} 
 \hline
 Name & $f(X)$ & $p_X$ & $T^o_f$/$T^p_f$ & $T^u_f$/$T^g_f$\\ [0.5ex] 
 \hline
 Facility Location & $\sum_{i \in V} \max_{k \in X} s_{ik}$ & $[\max_{k \in X} s_{ik}, i \in V]$ & $O(n^2)$ & $O(n)$\\ 
 \hline
 Saturated Coverage & $\sum_{i \in V} \min\{\sum_{j \in X} s_{ij}, \alpha_i\}$ & $[\sum_{j \in X} s_{ij}, i \in V]$ & $O(n^2)$ & $O(n)$\\ 
\hline
 Graph Cut & $\lambda \sum_{i \in V}\sum_{j \in X} s_{ij} - \sum_{i, j \in X} s_{ij}$ & $[\sum_{j \in X} s_{ij}, i \in V]$ & $O(n^2)$ & $O(n)$\\ 
\hline
 Feature Based & $\sum_{i \in \mathcal F} \psi(w_i(X))$ & $[w_i(X), i \in \mathcal F]$ & $O(n|\mathcal F|)$ & $O(|\mathcal F|)$ \\
 \hline
 Set Cover & $w(\cup_{i \in X} U_i)$ & $\cup_{i \in X} U_i$ & $O(n|U|$ & $|U|$\\
 \hline
  Clustered Set Cover & $\sum_{i = 1}^k w(\Gamma(X) \cap C_i)$ & $[\Gamma(X) \cap C_j, j \in 1, \cdots, k]$ & $O(n|U|$ & $|U|$\\
 \hline
 Prob. Set Cover & $\sum_{i \in \mathcal U} w_i[1 - \prod_{k \in X}(1 - p_{ik})]$ & $[\prod_{k \in X} (1 - p_{ik}), i \in \mathcal U]$ & $O(n|\mathcal U|)$ & $O(|\mathcal U|)$ \\ [1ex] 
 \hline
Spectral Functions & $\sum_{i = 1}^{|X|} \psi(\lambda_i(S_X))$ & SVD($S_X$) & $O(|X|^3)$ & $O(|X|^2)$\\
 \hline
 DPP & $\log\det(S_X))$ & SVD($S_X$) & $O(|X|^3)$ & $O(|X|^2)$\\
 \hline
 Dispersion Min & $\min_{k,l  \in X, k \neq l} d_{kl}$ & $\min_{k, l \in X, k \neq l} d_{kl}$ & $O(|X|^2)$ & $O(|X|)$\\
 \hline
 Dispersion Sum & $\sum_{k,l  \in X} d_{kl}$ & $[\sum_{k \in X} d_{kl}, l \in X]$ & $O(|X|^2)$ & $O(|X|)$\\
  \hline
 Dispersion Min-Sum& $\sum_{k \in X} \min_{l \in X} d_{kl}$ & $[\min_{k \in X} d_{kl}, l \in X]$ & $O(|X|^2)$ & $O(|X|)$\\
 \hline
\end{tabular}}
\caption{List of Functions used, with the precompute statistics $p_X$, gain evaluated using the precomputed statistics $p_X$ and finally $T^f_o = T^f_p$ as the cost of evaluation of the function without memoization and $T^f_u = T^f_g$ as the cost with memoization. It is easy to see that memoization saves an order of magnitude in computation. }
\end{center}
\end{table*}

\section{Precomputational Complexity Model}
Given a specific submodular function $f$, we denote the precomputed statistics as $p_X$. This data structure depends on the specific submodular function, and stores information about a given set $X$. The idea here is that given $p_X$, computing the gains $f(j | X, p_X)$ is often less costly compared to computing $f(j | X)$ by evaluating $f(X \cup j)$ and $f(X)$. This same idea holds for computing $f(j | X \backslash j, p_X)$. 
The simplest precomputed statistic, which applies to any submodular function in the oracle model, is storing $p_X = f(X)$. Evaluating the gains $f(j | X, p_X)$ just requires one oracle call as opposed to two, when computing $f(j | X)$. 
This simple trick can matter a lot since many of the algorithms require computing many gains (i.e., $f(j | X)$ for many $j \in V$). For example, given the statistic $p_X$, the modular upper bounds can be easily computed since they depend on $f(j | X), \forall j \notin X$ and $f(j | X \backslash j), \forall  j \in X$. Given a general submodular function in the value oracle model, this gives a factor $2$ speedup. Many real world submodular functions, however, have richer statistics, often enabling speedups up to a factor of $O(n)$ or even $O(n^2)$. The same idea holds in computing modular lower bounds and extreme points of the submodular polyhedron. In this case, we are given a permutation $\sigma$, and we compute a chain of gains $f(\sigma(i) | S^{\sigma}_{i-1}), \forall i \in V$. Often it is also much easier to update the statistic when adding an element to a set -- i.e., given a set $X$ and its statistic $p_X$, one can easily compute $p_{X \cup j}$ using $p_X$ (i.e., without needing to compute it from scratch). Similarly, we can ``downdate'' the statistics, i.e., given the statistics $p_X$, we would like to compute the statistics for $p_{X \backslash j}$.

Most submodular function optimization algorithms either use the modular upper bounds (in which case, they compute $f(j | X)$ repeatedly), or they compute extreme points of the submodular polyhedron (in the form of subgradients, or via the \lovasz{} extension), or they greedily add or remove elements. These algorithms easily admit our memoization framework.

Denote $\mathbf{T^o_f}$ as the oracle complexity of the submodular function, a quantity which most existing algorithms use for their analysis. Define $\mathbf{T^g_f}$ as the complexity of evaluating the gains, given the precomputed statistics. This depends in general on whether we are adding or removing the item (i.e., for $f(j | X, p_X)$ or $f(j
| X \backslash j, p_X)$). For submodular functions we consider here, however, both complexities are the same. Also denote $\mathbf{T^u_f}$ as the complexity of updating the precomputed statistics. Again, the complexity of updating or downdating could be different in general, but they are the same for the submodular functions we consider here. Finally, denote $\mathbf{T^p_f}$ as the complexity of computing the precomputed statistic for a set $X$ from scratch.

Under the value oracle model, the complexity of evaluating the modular upper bounds is $O(nT^o_f)$, since the modular upper bound requires $n$ oracle queries. In the precomputational model, the complexity is $O(nT^g_f + T^p_f)$, since we would need to compute the precomputed statistic $p_X$ (in the worst case) from scratch, and then compute $f(j | X, p_X)$. For most submodular functions, $T^p_f$ is roughly the same order as $T^o_f$, but $T^g_f$ is often at least a factor $n$ cheaper. Similarly, the complexity of computing the modular lower bound is $O(nT^o_f)$ in the value oracle model, but is $O(n[T^g_f + T^u_f])$ in the precomputational model. Again for almost all submodular functions, $T^g_f$ and $T^u_f$ are at least a factor $n$ cheaper than $T^o_f$. Hence the precomputational model provides significant speedups to the algorithms, which can be very important in large scale machine learning problems.

We make the ideas above more concrete in the following sections, by first explicitly defining $p_X$ and the procedure for updating $p_X$, for several classes of submodular functions which occur in applications. Table 1 summarizes the precompute statistics $p_f(X)$, the complexity $T_f^o$/$T_f^p$ and $T_f^u$/$T_f^g$ for different functions $f$.

\subsection{Graph Based Submodular Functions}
\textbf{Facility Location Functions: }Given a similarity matrix $\{s_{ij}\}_{i, j \in V}$ the facility location function is $f(X) = \sum_{i \in V} \max_{j \in X} s_{ij}$. This function has successfully been used in summarization and data subset selection~\cite{linbudget, lin2009graph}. It is easy to check that the oracle complexity $T_f^o = O(n^2)$. The precomputed statistics in this case is $p_X[i]$ is the pair of the largest and second largest value of $s_{ij}: i \in V, j \in X$. Given these statistics however computing the gains are much easier, since $f(k | X, p_X) = \sum_{i \in V} \max\{0, s_{ik} - \max_{j \in X} s_{ij}\}$, and hence $T^g_f = O(n)$. We can similarly efficiently compute the gains $f(k | X \backslash k, p_X)$ in $O(n)$. Moreover, updating the statistics is also easy, since $\max_{j \in X \cup k} s_{ij} = \max\{s_{ik}, \max_{j \in X} s_{ij}\}$ -- we can similarly update the second largest. Hence $T^u_f = O(n)$. Moreover, computing $p_X$ is the same complexity of computing $f$, and $T_f^p = T_f^o$. We can easily extend this idea to the top-$k$ facility location function, where instead of taking the $\max$, we can take the top $k$ similarities. In that case, the precomputed statistics would be a matrix $p_X[i, l], i \in V, l \in \{1, \cdots, k\}$ as the $l$th largest value of $s_{ij}$ for a given $i$. It is easy to see that we obtain an $O(n)$ speedup in this case as well.

\textbf{Saturated Coverage functions: } The saturated coverage function, $f(X) = \sum_{i \in V} \min\{\sum_{j \in X} s_{ij}, \alpha_i\}$, has successfully been used in document summarization~\cite{linacl}. The oracle complexity of this function is also $T_f^o = O(n^2)$. A natural choice of the precomputed statistics in this case is $p_X[i] = \sum_{j \in X} s_{ij}, i \in V$. Given this, it is easy to compute $f(j | X, p_X)$ in $O(n)$ time. Moreover, updating $p_X$ can also be done directly in $O(n)$, since $p_{X \cup k}[i] = p_X[i] + s_{ik}$. Hence in this case also, $T^g_f = T^u_f = O(n)$ and $T^p_f = O(n^2)$. 

\textbf{Graph Cut like functions: }This class of functions have been used extensively in both summarization problems, while modeling coverage and diversity~\cite{lin2009graph, linbudget}, as well in image segmentation and denoising, by capturing cooperation~\cite{frbach1}. We can denote the general class as $f(X) = \lambda \sum_{i \in V}\sum_{j \in X} s_{ij} - \sum_{i, j \in X} s_{ij}$. Setting $\lambda = 1$ is the standard graph cut, while $\lambda = 0$ gives the redundancy penalty~\cite{lin2009graph}. Again, the natural choice of $p_X[i] = \sum_{j \in X} s_{ij}$. Though $T_f^o = O(n^2)$, $T_f^g = T^u_f$ are both $O(n)$. Similarly, $T_f^p = O(n^2)$.\looseness-1

\subsection{Coverage Functions}
\textbf{Set Covers and Neighborhood Functions:} This is another important function,
capturing the notion of coverage~\cite{lin2012submodularity} in maximization problems. This function also captures notions of complexity (like the size of the vocabulary in a speech corpus) in minimization problems~\cite{lin2011optimal}. Given a
set of sets $\{\mathcal S_1, \cdots, S_n\}$ and the universe $\mathcal
U = \cup_i \mathcal S_i$, define $f(X) = w(\cup_{i \in X} \mathcal
S_i)$, where $w_j$ denotes the weight of item $j \in \mathcal U$. This
setup can alternatively be expressed via a neighborhood function
$\Gamma: 2^V \rightarrow 2^{\mathcal U}$ such that $\Gamma(X) =
\cup_{i \in X} \mathcal S_i$. The oracle complexity of the function in this case is $T^o_f = O(|\mathcal U|)$. The precomputed statistics here is a vector $p_X \in \mathbb{R}^{\mathcal U}$ which stores the number of times $u \in \mathcal U$ appears in the sets $\{\mathcal S_i\}_{i \in X}$. Then, $f(j | X, p_X)  = |\{u \in \mathcal S_j: p_X[u] = 0\}|$, and $f(j | X \backslash j, p_X) = |\{u \in \mathcal S_j: p_X[u] = 1\}|$. Similarly, $p_{X \cup j}$ can be updated as $p_{X \cup j}[u] = p_X[u] + 1$, for $u \in \mathcal S_j$. It is easy to see that both $T_f^u = T_f^g = O(1)$, and $T^p_f = T^u_f$ (assuming $|S_j|$ is a constant, which is often the case).

\textbf{Clustered Set Cover: } We can generalize this idea to clustered set cover functions, often used as confusability functions in corpus selection~\cite{lin2011optimal}. This is defined as $f(X) = \sum_{i = 1}^k w(\Gamma(X) \cap C_i)$, where $C_1, C_2, \cdots C_k \subseteq \mathcal U$ are clusters. The precomputed statistics here is a vector of sets. For each $j = 1, \cdots, k$, $p_X[j] = \Gamma(X) \cap C_j$.

\textbf{Probabilistic Coverage Functions: }Another generalization of the set cover function, which has been used in a
number of models for summarization problems~\cite{el2009turning}. This
provides a probabilistic notion to the set cover function, and is defined as $f(X) = \sum_{i \in \mathcal U} w_i [1 - \prod_{j \in X} (1
- p_{ij})]$. The complexity of evaluating this function $T^o_f = O(n|\mathcal U|)$. The precomputed statistics in this case is $p_X[i] = \prod_{j \in X} (1
- p_{ij})$. Note that here, both $T^u_f = T^g_f = O(|\mathcal U|)$, thereby providing a factor $n$ speedup. Similarly, $T^p_f = T^u_f$.

\subsection{Feature Based Functions and Clustered Concave over Modular Functions}
Another class of submodular functions are sums of concave over modular functions. They appear in maximization problems as feature based functions, defined as $f(X) = \sum_{e \in \mathcal F} \psi(m_{e}(X))$, and have been used in data subset selection applications~\cite{wei2014submodular}.  $m_e(j)$ captures how much item $j$ covers feature $e \in \mathcal F$. Another related function is $f(X) = \sum_{j = 1}^k \psi(m_j(X \cap C_j))$, where $C_1, C_2, \cdots, C_k$ are clusters of similar items in the ground set $V$. This function simultaneously captures diversity in maximization problems~\cite{linacl}, and notions of cooperation in minimization problems~\cite{jegelka2011-nonsubmod-vision, rkiyeruai2012}. The complexity of evaluating these functions is $O(n |\mathcal F|)$ and $O(nk)$ respectively. A natural choice of the precomputed statistics is $p_X[e] = m_e(X)$, in the case of feature based functions, and $p_X[j] = m_j(X \cap C_j)$ for clustered concave over modular functions. Again, it is easy to see that $T^u_f = T^g_f$ are $O(|\mathcal F|)$ and $O(k)$ respectively, thus saving a factor of $n$. Moreover, in this case we also have that $T^p_f = T^o_f$.\looseness-1

\subsection{Spectral Submodular Functions}
Another rich class of submodular functions, defined as $f(X) = \sum_{i = 1}^{|X|} \psi(\lambda_i(S_X))$, where $S$ is a PSD matrix, $S_X$ represents the principal submatrix formed by the rows and columns corresponding to $X$, and $\psi$ is a concave function. This function is submodular for a large class of concave functions~\cite{friedland2011submodular}. This class of spectral regularizers has been shown to promote diversity~\cite{das2012selecting}, and includes as special cases the $\log det$ function, which occurs in the context of determinantal point processes~\cite{kulesza2012determinantal}, since $f(X) = \log \det(S_X) = \sum_{i = 1}^{|X|} \log \lambda_i(S_X)$. Another example of this function is $f(X) = \sum_{i = 1}^{|X|}\sqrt{\lambda_i(S_X)}$. Evaluating this class of functions is $T^o_f = O(n^3)$, since we need to perform the eigenvalue decomposition. A natural choice of the precomputed $p_X$ statistic here is the eigenvalue decomposition of $S_X$. Using the result from~\cite{gu1994stable}, it is possible to update (or downdate) the eigenvalue decomposition of $S_X$ to $S_{X \cup j}$ (or $S_{X \backslash J}$) in $O(n^2)$ computations given the eigenvalue decomposition of $S_X$ (note that the result of \cite{gu1994stable} is in terms adding a single row and column to $S_X$. However, converting $S_X$ to $S_{X \cup j}$ is the same as adding a row, followed by a column, and hence two updates.)\looseness-1

\subsection{Dispersion Functions}
Denote $d_{ij}$ as a distance measure between objects $i$ and $j$. Define
the `Dispersion Min function as $f(X) = \min_{i, j \in X} d_{ij}$. This function is not submodular, but can be efficiently optimized via a greedy algorithm~\cite{dasgupta2013summarization}. It is easy to see that maximizing this function involves obtaining a subset with maximal minimum pairwise distance, thereby ensuring a diverse subset. Similarly, we can define two more variants. One is
the supermodular Dispersion Sum, defined as $f(X) = \sum_{i, j \in X} d_{ij}$. Another
is Dispersion Min-Sum, a combination of two forms, defined as $f(X) = \sum_{i \in X} \min_{j \in X} d_{ij}$. This function is submodular~\cite{chakraborty2015adaptive}. 

\subsection{Mutual Information and Entropy}
This is another class of functions, used often for feature subset
selection~\cite{rkiyeruai2012}.  The entropy function $f(A) = H(X_A)$
is submodular, and while mutual information $f(A) = I(X_A; C)$ is
always a difference of submodular functions
\cite{narasimhanbilmes,rkiyeruai2012}, it is also sometimes submodular
under some assumptions.  While both these functions require
exponential complexity to evaluate, they can be estimated easily from
data via a single sweep, and using techniques like Laplacian
smoothing~\cite{rkiyeruai2012} -- the computational complexity being
$O(n|\mathcal D|)$, where $|\mathcal D|$ is the size of the training
data. The mutual information and entropy estimates also amenable to
precomputation, since we precompute the data tables for the given set
of feature $X_A$. Adding and removing features to this corresponds to
further dividing tables, which can be done in $O(|\mathcal D|)$
complexity.

\subsection{Influence Maximization}
A number of models for influence maximization have been shown to be related to submodular maximization~\cite{kkt03}. Evaluating the objective function, however, requires MCMC simulations, which is quite expensive. We can define precomputed statistics however, which can significantly speedup the greedy algorithms~\cite{goyal2011simpath}. In particular, they avoid the expensive MCMC by enumerating simple paths, and relying on memoization and look ahead optimization~\cite{goyal2011simpath}.

\subsection{Mixtures of Submodular Functions}
Often it is desirable to consider not just one submodular function, but a mixture of many submodular bases functions~\cite{lin2012submodularity}. In particular, we often express $f(X) = \sum_{i = 1}^m w_i f_i(X)$, where $f_i$'s are bases submodular functions (like one of the submodular functions above), and $w_i$'s are weights. Assuming the submodular functions $f_i$'s have precomputed statistics, $p^{f_i}_X$, the gain $f(j | X, p^f_X)$ is exactly $\sum_{i = 1}^m w_i f_i(j | X, p^{f_i}_X)$. Moreover, to update the statistics $p^f_X$, we simply update the individual $p^{f_i}_X$.\looseness-1

\subsection{Deep Submodular Functions}
Almost all of the above classes of submodular functions are subsumed
by Deep Submodular Functions~\cite{bilmes2017deep}, defined as nested
sums of concave over fewer-layer deep submodular functions. The
general form of Deep SFs are:
\begin{align*}
f(X) = \sum_{i_1 \in \mathcal F_1} w^1_{i_1} \psi_1(\cdots \psi_{k-1}(\sum_{i_{k-1} \in \mathcal F_k} w^k_{i^k} \psi_k (m^{k}_{i_k}(X)))) 
\end{align*}
For simplicity we consider the case with $k = 2$, i.e., a two layer function. The complexity of evaluating $f$ is $|\mathcal F_1||\mathcal F_2|n$. Similar to a feature based function, the precomputed statistics here is $p_X = m^2_{i_2}(X), \forall i_2 \in \mathcal F_2$. The complexity of updating the precomputed statistics is $T_f^u = |\mathcal F_2|$ while the complexity of computing the gain $T_f^g = |\mathcal F_1||\mathcal F_2|$. Both these quantities are a factor $n$ less expensive compared to $T_f^o$.

\begin{table*}
\begin{center}
\small{
 \begin{tabular}{|| c | c | c  ||} 
 \hline
 Algorithm & Value Oracle Model & Precomputational Model\\ [0.5ex] 
 \hline
 Computing Supergradients & $2nT_f^o$ & $nT_f^g + T_f^p$ \\ \hline
 Computing Subgradients/Extreme Points & $nT_f^o$ & $n(T_f^g + T_f^u)$ \\ \hline
 Min-Max Alg. Framework & $\tilde{O}(n T_f^o)$ & $\tilde{O}(n (T_f^g + T_f^u))$ \\ \hline
 Greedy Algorithm & $O(nk T_f^o)$ & $O(nk (T_f^g + T_f^u))$ \\ \hline
 Lazy Greedy Algorithm & $O(k n_R T_f^o)$ & $O(k n_R (T_f^g + T_f^u))$ \\ \hline
  Lazier than Lazy Greedy Algorithm & $O(n\log(1/\epsilon) T_f^o)$ & $O(n\log(1/\epsilon) (T_f^g + T_f^u))$ \\ \hline
    Sieve Streaming Algorithm & $O(n\log(k/\epsilon) T_f^o)$ & $O(n\log(k/\epsilon) (T_f^g + T_f^u))$ \\ \hline
    Distributed Greedy Algorithm & $O([nk/m + mk^2] T_f^o)$ & $O([nk/m + mk](T_f^g + T_f^u))$ \\ \hline
      Local Search Algorithm & $O(n^3 \log n T_f^o/\epsilon)$ & $O(n^3 \log n (T_f^g + T_f^u)/\epsilon)$ \\ \hline
    Bi-directional Greedy Algorithm & $O(n T_f^o)$ & $O(n(T_f^g + T_f^u))$ \\ \hline
    Randomized Greedy Algorithm & $O(nk T_f^o)$ & $O(nk (T_f^g + T_f^u))$ \\ \hline
    Minimum Point Algorithm & $O([n^5 T_f^o + n^7]F^2)$ & $O([n^5(T_f^g + T_f^u) + n^7]F^2)$ \\ \hline
    Lovasz Extension based Algorithm & $O(T_f^o/\epsilon^2)$ & $O((T_f^g + T_f^u)/\epsilon^2)$ \\ \hline
    Minorization-Maximization Algorithm & $\tilde{O}(nT_f^o)$ & $\tilde{O}(T_f^p + nT_f^g)$ \\ \hline
\end{tabular}}
\caption{List of Submodular Optimization Algorithms, and their complexity under the value oracle model and the Precomputational Model. See text for more details on the quantities in this Table}
\end{center}
\end{table*}

\section{Algorithms for Submodular Optimization}
We now investigate several known algorithms for submodular optimization problems, and show how we can easily integrate the precomputation idea into them. In almost all cases, we shall see that this entails only a few additional lines of code, while providing significant speedups in applications. Table 2 summarizes the complexity of various submodular optimization algorithms with the precomputational model and value oracle model.\looseness-1

\subsection{Computing Subgradients, Supergradients, and extreme points of Submodular Polyhedron}
Most submodular optimization algorithms either rely on computing subgradients, supergradients or some extreme points of the submodular polyhedron. So we first compare the complexity of computing these quantities under the Precomputational Model and the Value Oracle Model. Recall that computing a supergradient of $X$ requires computing $f(j | X)$ for every $j \notin X$ (or equivalently computing $f(j | X \backslash j)$ for $j \in X$). The complexity of doing this in the Value Oracle Model is $O(nT_f^o)$. Under the Precomputational Model, the complexity if $O(T_f^p + nT_f^g)$ since we can compute the precompute statistics $p_X$ and using that, evaluate the gains $f(j | X, p_X)$. This is a factor $n$ speedup since in all cases, $T_f^g$ is a factor $n$ cheaper compared to $T_f^0$ (see Table 1). Similarly, computing a subgradient (or equivalently computing an extreme point of the submodular polyhedron) requires forming a chain of sets $\emptyset = X_0 \subseteq X_1 \subseteq \cdots \subseteq X_n = V$, and computing $f(x_i | X_i)$ where $x_i = X_{i+1} \backslash X_i$. Computing the subgradient (or extreme point) using the Value Oracle Model is $O(nT_f^o)$. Using the precompute statistics, we can at every step use the precompute statistics from $X_i$ to compute $f(x_i | X_i)$ and then update the precompute statistics. Since this is a greedy algorithm, the complexity of this is $O(n[T_f^g + T_f^u]$. Again, from Table 1, it is evident that we can achieve a speedup at least of a factor of $n$.
 
\subsection{Submodular Maximization}
Submodular maximization is particularly important in applications like summarization~\cite{linacl, linbudget}, data subset selection~\cite{wei2014submodular} etc. where we want to find diverse and relevant subsets.
A large class of existing submodular maximization algorithms can be expressed via a common minorization-maximization framework~\cite{rkiyersemiframework2013} --  an iterative procedure which optimizes the modular lower bound $h^f_{\sigma^t}(X)$ (which is tight at $X^t$). This algorithm essentially chooses a sequence of orderings $\sigma^t$, each of which is tight with respect to the set $X^t$, and different known algorithms use different orderings. The simplest algorithm is to just choose a random subgradient (or ordering) $\sigma^t$ at every iteration. We can compute the subgradients $h^f_{\sigma^t}$ using memoization -- start with the empty set, and compute $f(\sigma^t(i) | S^{\sigma^t}_{i-1}, p_{S^{\sigma^t}_{i-1}})$, and update $p_{S^{\sigma^t}_i}$, for $i = 1, 2, \cdots, n$. The complexity of this algorithm using the precompute statistics is $\tilde{O}(n[T_f^g + T_f^u])$ -- where $\tilde{O}$ hides the complexity of the outer loop which is weakly polynomial~\cite{rkiyersemiframework2013} (in practice, it is a constant). With the value oracle model, the complexity of minorize-maximize is $\tilde{O}(n[T_f^o])$.

While this simple algorithm provides guarantees, one gets much tighter bounds with more intelligent choices of subgradients. Below we look at a few such algorithms for various variants of submodular maximization.\looseness-1

\textbf{Greedy and Lazy Greedy Algorithm: } This is a common algorithmic idea, which provides constant factor guarantees for a large class of monotone submodular maximization problems, under cardinality, knapsack and matroid constraints~\cite{nemhauser1978}. Starting with $X^0 = \emptyset$, we sequentially update $X^{t+1} = \argmax_{j \in V \backslash X^t} f(j | X^t)$. The complexity of this algorithm is $O(nk T_f^o)$. We can easily integrate precomputation into this, by setting the update rule to $X^{t+1} = \argmax_{j \in V \backslash X^t} f(j | X^t, p_{X^t})$, and updating $p_{X^{t+1}}$. The complexity then is essentially $O(nk [T_f^u + T_f^g])$. Thanks to submodularity, however, we can significantly accelerate this algorithm, to what is known as \emph{the lazy greedy algorithm}~\cite{minoux1978accelerated}. The idea is that instead of recomputing $f(j | X^t), \forall j \notin ^t$, we maintain a priority queue of sorted gains $\rho(j), \forall j \in V$. Initially $\rho(j)$ is set to $f(j), \forall j \in V$. The algorithm selects an element $j \notin X^t$, if $\rho(j) \geq f(j | X^t)$, we add $j$ to $X^t$ (thank to submodularity). If $\rho(j) \leq f(j | X^t)$, we update $\rho(j)$ to $f(j | X^t)$ and resort the priority queue. The complexity of this algorithm is roughly  $O(k n_R T_f^o)$, where $n_R$ is the average number of resorts in each iteration. Note that $n_R \leq n$, while in practice, it is a constant thus offering almost a factor $n$ speedup compared to the simple greedy algorithm. We can use the notion of precompute here too, and use $f(j | X^t, p_{X^t})$ in place of the gain $f(j | X^t)$. Note that additionally, whenever we add an element $j$ to $X^t$, we also need to update the precomputed statistic. The resulting complexity in precomputational model is $O(k n_R [T^g_f + T^u_f])$, again providing a factor $n$ speedup.\looseness-1

\textbf{Lazier than Lazy greedy Algorithm: } While the lazy greedy algorithm above runs much faster in practice, in the worst case, the complexity is the same as the na\"ive greedy algorithm. The Lazier than Lazy greedy algorithm~\cite{mirzasoleiman2015lazier} attempts to obtain an approximation guarantee of $1 - 1/e - \epsilon$ in $O(n \log(1/\epsilon)$ function evaluations rather than $O(nk)$ from the greedy (or lazy greedy) algorithm. The idea is to select a random set $R$ of size $n/k \log(1/\epsilon)$ at select the element with the largest gain of adding the element to the current set. We run this until our chosen subset has $k$ elements (in the cardinality constrained case). Since we evaluate the gains $f(j | X), j \in R \subseteq V \backslash X$. We start with the $X = \emptyset$, and at every step we can compute $f(j | X, p_X)$ using the precompute statistics, and after choosing the best element, add that element to $X$ and update $p_X$. The complexity of this algorithm under the precompute model is $O(n \log(1/\epsilon) [T_f^g + T_f^u])$, while using the value oracle model, the complexity is $O[n \log(1/\epsilon) T_f^o]$.

\textbf{Distributed Greedy Algorithm: } The Distributed Greedy algorithm~\cite{mirzasoleiman13distributed} attempts to extend the Greedy algorithm to a setting where all the data cannot fit into memory. The basic idea is we have $m$ machines, where we equally partition the data into, i.e., $V_1, \cdots, V_m$. The distributed greedy algorithm then runs a greedy algorithm on each of the partitions to obtain $k$ elements, followed by a second round of greedy on the $mk$ elements to obtain $k$ elements. It is easy to see that the complexity of this in the value oracle model is $O([nk/m + mk^2]T_f^o)$. Moreover, since we run a two-stage greedy algorithm, the memoization can be used exactly like the memoization discussed above.

\textbf{Sieve Streaming Algorithm: } The Sieve streaming algorithm basically performs submodular maximization in the streaming setting~\cite{badanidiyuru2014streaming}. The idea of the algorithm is to maintain a set of thresholds in a set $O$, and corresponding to each threshold, maintain a set $S$. It simultaneously grows these sets depending on constraints (see~\cite{badanidiyuru2014streaming} for details on the algorithm). The way we can incorporate memoization is by storing $|O|$ copies of submodular functions, with their own precompute statistics, which are updated through the process of the algorithm. Given these precompute statistics, we can easily compute the gains $f(e_i | S_i)$ for each of the sets maintained by the algorithms, and once the elements are added to the sets, we can update the precompute statistics. The complexity analysis (shown in Table 1) follows directly from the results in~\cite{badanidiyuru2014streaming} and the precompute statistics (note that each subset is simultaneously updated through the course of the algorithm in a greedy manner).  

\textbf{Local Search Algorithm: } This algorithm for unconstrained submodular maximization (USM), essentially runs multiple rounds of the greedy algorithm~\cite{janvondrak}, and provides a $1/3$ approximation for USM. In particular, we start with $X^0 = \emptyset$, and run the forward greedy algorithm until we can no longer add elements, followed by the reverse greedy of removing elements. We continue this procedure until we converge to a local optimum. The forward greedy algorithm is essentially the same algorithm as above, while in the reverse greedy case, we remove elements with the smallest value of $f(j | X^t \backslash j)$. We can use the precompute ideas and use $f(j | X^t \backslash j, p_{X^t})$ (in this case, we downdate the statistics after removing the elements).  The complexity of the algorithm follows from the complexity of the local search from~\cite{janvondrak} -- see Table 2 for the complete expression. Again, we simply need to update and downdate the precompute statistics, and given these statistics, compute the gains of adding and removing the elements. For all the functions we consider, the complexity of adding/removing elements (correspondingly updating the downdating the precompute statistics) is the same, the complexity analysis follows.

\textbf{Bidirectional Greedy Algorithm: } The bidirectional greedy algorithm~\cite{feldman2012optimal} provides the tight $1/2$ approximation for USM. Surprisingly, this is a very simple linear time algorithm. This algorithm maintains two sets $A$ and $B$ (initially set to $A = \emptyset$ and $B = V$) which increases and decreases respectively in the course of the algorithm, and depends on an initial ordering $\pi$. Then from $i = 1, 2, \cdots, n$, we either add $\pi(i)$ to $A$ or remove $\pi(i)$ from $B$, depending on which of the gains is larger. The complexity of this algorithm is $O(n T_f^o)$. We can use precomputation here, by storing two sets of statistics (one for the set $A$, and another for $B$) -- in practice, we can achieve this by maintaining two submodular functions which each store their statistics. As $A$ grows, we update its statistics and similarly downdate $B$'s statistics as it shrinks. The complexity with precomputation is $O(n [T^g_f + T^u_f])$, which is in practice a factor $n$ faster.\looseness-1

\textbf{Randomized Greedy Algorithm: } 
The randomized greedy algorithm provides an efficient algorithmic framework for cardinality constrained non-monotone submodular functions. The idea of this algorithm is very similar to the bidirectional greedy above, except that instead of choosing the best gain $f(j | X^t)$, we choose at random, one of the top $k$ gains (where $k$ is the given cardinality constraint). The complexity of this algorithm is $O(nk T^o_f)$. Similar to the greedy algorithm, we can incorporate the precomputation by using $f(j | X^t, p_{X^t})$, and updating $p_{X^t}$ when we add the new element. The complexity of this is $O(nk [T^u_f + T^g_f])$.\looseness-1

\subsection{Submodular Minimization}
Submodular minimization comes up in applications where we want to minimize cooperative costs and complexity measures, like image segmentation~\cite{frbach1, jegelka2011-nonsubmod-vision}, and limited vocabulary corpus selection~\cite{lin2011optimal}.
We investigate three important submodular minimization algorithms, and show how the notions of precomputation yields significant computational gains.

\textbf{The minimum norm point algorithm: } This is one of the most practical algorithm available for general purpose submodular minimization~\cite{fujishige2011submodular, frbach1, lin2012submodularity}. This algorithm solves quadratic programming problem, which is equivalent to of the discrete minimization problem~\cite{fujishige2005submodular}, and uses the Wolfe algorithm~\cite{frank1956algorithm}. One of the most important step in this algorithm (and the only step which requires oracle access to the submodular function) is the greedy algorithm for solving a linear programming problem $\max_{x \in \mathcal P_f} \langle x, \hat{x} \rangle$.  As discussed in section~\ref{basicideas}, solving this problem requires computing the subgradient according to the ordering $\sigma_{\hat{x}}$, and we can use precomputations to efficiently find these subgradients, thereby providing significant speedups in practice. While in practice the Minimum Norm Point algorithm is the fastest (compared to the other combinatorial algorithms), the worst case complexity is still high order polynomial~\cite{chakrabarty2014provable}. The worst case complexity with and without memoization is in Table 2. Having said that, we demonstrate in our experimental section, that memoization can provide significant compute gains.

\textbf{\lovasz{} extension based: } Another class of algorithms~\cite{frbach1, iyer2014monotone}, is based on relaxing the discrete optimization problem to a continuous one, via the \lovasz{} extension. This procedure, moreover, works for a large class of constrained problems, and uses convex optimization techniques~\cite{frbach1}. The precomputational ideas apply in these cases too, since we can both compute the \lovasz{} extension (which requires solving the linear program over the submodular polyhedron) and its subgradient (which is the same as the subgradient of the submodular function) very efficiently.

\textbf{Majorization-Minimization: } This is a discrete gradient based framework which applies to a large class of constrained submodular minimization problems~\cite{narasimhanbilmes,rkiyeruai2012,rkiyersemiframework2013}. This is an iterative procedure, which starts with $X^0 = \emptyset$, and minimizes the modular upper bounds $m^f_{X^t}$ as a proxy to $f$. Each step of this algorithm is a linear cost problem, which is easy for many combinatorial constraints.  Thanks to the nature of this algorithm, we are guaranteed improvement at every iteration. Moreover, this algorithm also admits guarantees, and works very well in practice~\cite{rkiyersemiframework2013}. Moreover, as we saw earlier, we can efficiently compute the modular upper bounds using precomputations. The complexity under this model is $O(nT^g_f + T^p_f)$, which is in general a factor $n$ faster than the oracle model $O(n T^o_f)$.

\subsection{Optimization with Submodular Constraints}
Another class of optimization problems related to submodular functions, are ones where submodular functions appear as upper or lower bound constraints. Two general problem classes here are~\cite{nipssubcons2013} (a) SCSC: $\min\{f(X) | g(X) \geq c\}$, and (b) SCSK: $\max\{g(X) | f(X) \leq b\}$. This class of problems comes in applications where we simultaneously want to maximize one submodular function, while minimizing another. 
This generalizes a number of useful problems, including for example, the submodular set cover~\cite{wolsey1982analysis} and the submodular knapsack~\cite{nemhauser1978}, which are instances of SCSC and SCSK respectively, when $f(X) = w(X)$ is a modular function and $g(X)$ is submodular function. The same lazy greedy algorithm actually works for the submodular set cover and the submodular knapsack problem, and hence precomputation directly applies in this case. The general case of SCSC and SCSK, can be handled by replacing $f$ by its modular upper bound and iteratively solving submodular set cover and submodular knapsack respectively~\cite{nipssubcons2013}. Moreover, since the modular upper bound computation and the greedy algorithm can be both done efficiently via precomputations, we can obtain significant speedups for these problems in large scale problems.

\subsection{Difference of Submodular Functions}
A very common and general optimization problem involves minimizing the difference between submodular functions $\min_{X \subseteq V} f(X) - g(X)$, and comes up in several machine learning applications including feature subset selection, and graphical model inference~\cite{narasimhanbilmes}. A common class of heuristics for this problem is the submodular-supermodular procedure, and its variants~\cite{narasimhanbilmes,rkiyeruai2012}. They are essentially majorization-minimization based iterative procedures starting with $X^0 = \emptyset$, and iteratively to replace either $f$ by its modular upper bound $m^f_{X^t}$, and $g$ by its modular lower bound $h^g_{X^t}$, or both. At every iteration, the resulting problem is either submodular minimization, maximization or modular minimization~\cite{narasimhanbilmes,rkiyeruai2012}. Since both the upper and lower bounds can be efficiently computed, and the algorithms for submodular minimization and maximization are efficient, thanks to precomputations, we can achieve substantial speedups in applications.

\section{Experiments} 
In this section, we compare the performance of different submodular optimization problems (functions and algorithms). We study two specific applications: 1) speech data subset selection~\cite{wei2013using}, and 2) low complexity speech corpus creation~\cite{liu2015svitchboard}.  We divide the sections below by the different submodular optimization problems. We denote the precompute model as PM and the value oracle model as VO. 

\textbf{Computing Sub- and Super-gradients:}
We first compare the running time of computing the sub and super gradients for the speech data subset selection problem~\cite{wei2013using} on TIMIT. In this case, $n = 4620$. We compare the running time of three classes of submodular functions: Facility Location, Feature Based, and the Complexity function from~\cite{liu2015svitchboard} (which is a set cover function). The Results (Table 3) shows substantial gains using memoization compared to the VO model for all three classes of functions. We also compute the sub and supergradients for large scale problems ($|V| = 200000$ from Switchboard~\cite{liu2015svitchboard}. In this case, the subgradient computation takes 2.8 seconds for the Feature based function and 0.7 seconds for the Complexity function, while the supergradient compute takes 3.09 seconds for the Feat based and 1.14 sec for the Complexity Function (under the PM model). With the VO models, it would have taken around 7 days to compute these. 
\begin{table}[]
    \centering
	\small{
    \begin{tabular}{|c|c|c|c|c|}
        \hline
         & \multicolumn{2}{|c|}{Subgrad} & \multicolumn{2}{|c|}{Supergrad} \\ \hline
         Function & VO & PM & VO & PM \\ \hline
         Fac Loc & 433.4 & 0.39 & 505.7 & 0.49 \\ \hline
         Feat Based & 3.78 & 0.019 & 14.8 & 0.021 \\ \hline
         Complexity & 28.9 & 0.02 & 48.2 & 0.03 \\ \hline
    \end{tabular}}
    \caption{Computation of Sub and Super gradients under the VO and PM models}
    \label{tab:my_label}
\end{table}

\begin{table}
    \centering
    \scriptsize{
    \begin{tabular}{ |c|c|c|c|c|c|c|}
        \hline
         & \multicolumn{3}{c|}{\textbf{PreCompute Model}} & \multicolumn{3}{c|}{\textbf{Value Oracle Model}} \\
        \hline
        \textbf{Function} & \textbf{5\%} & \textbf{15\%} & \textbf{30\%} & \textbf{5\%} & \textbf{15\%} & \textbf{30\%}  \\
         \hline
        Fac Loc & 0.34 & 0.4 & 0.71 & 48 & 168 & 270  \\
         \hline
        Sat Cov & 0.36 & 0.64 & 0.92 & 55 & 177 & 301  \\
        \hline
         Gr Cut & 0.39 & 0.52 & 0.82 & 41 & 161 & 355  \\
        \hline
        Feat B & 0.16 & 0.21 & 0.32 & 9 & 16 & 21  \\
        \hline
        Set Cov & 0.21 & 0.31 & 0.41 & 5 & 16 & 31  \\
        \hline
        PSC & 0.11 & 0.37 & 0.42 & 7 & 19 & 35  \\
        \hline
        DM & 0.11 & 0.61 & 0.82 & 21 & 125 & 221  \\
                \hline
        DS & 0.21 & 0.63 & 0.89 & 41 & 134 & 246  \\
        \hline
    \end{tabular}}
    \caption{Timing results in seconds submodular maximization}
    \label{tab:my_label}
\end{table}

\textbf{Submodular Maximization}
Next we compare the different functions on submodular maximization with the lazy greedy algorithm. Again, we see substantial speedups using the precompute model across the board for different submodular functions (Table 4). We also compare the Feature Based and Set Cover function for large scale submodular optimization $|V| = 200000$. The FB function takes 16 seconds, while the Set Cover function takes about 27 seconds for running the Lazy greedy algorithm. 

\textbf{Submodular Minimization}
We next compare the complexity of unconstrained and constrained submodular minimization. We define $f(X) = c(X) - \lambda |X|$, where $c(X)$ is the complexity function from~\cite{liu2015svitchboard}. We use the TIMIT dataset ($|V| = 4620$. For unconstrained minimization, the minimum norm point algorithm takes about 7.2 seconds with PM, while under the VO Model, it takes around 2000 seconds. For larger scale problem ($n = 200000$), the MN algorithm takes around 88 seconds under the PM Model (the VO model will take several days to complete). Consider next the problem of constrained submodular minimization under cardinality constraints (minimizing $c(X)$ subject to a cardinality constraint). We use the Majorization-Minimization algorithm here~\cite{rkiyersemiframework2013}. Under PM, majorization-minimization takes around 0.13 seconds for a budget of 10\%, while the value oracle model, the time is 163 seconds. For the large scale version of this problem with Switchboard ($n = 200000$), MMin takes around 23 seconds with the PM Model.

\section{Conclusions}
This paper introduces the idea of precompute Statistics and Memoization for Submodular Optimization. We show how several real world submodular functions admit natural precompute statistics, and how we can integrate this idea into a large family of algorithms for submodular maximization, minimization and other forms of constrained submodular programs. We empirically demonstrate the utility of our Memoization framework on several large scale problems.

This material is based upon work supported by the National Science
Foundation under Grant No. (IIS-1162606), and a Google and a Microsoft
award. This work was supported in part by the CONIX Research Center,
one of six centers in JUMP, a Semiconductor Research Corporation (SRC)
program sponsored by DARPA.

\bibliographystyle{abbrv}
\bibliography{submod}

\begin{thebibliography}{10}

\bibitem{asadpour2008stochastic}
A.~Asadpour, H.~Nazerzadeh, and A.~Saberi.
\newblock Stochastic submodular maximization.
\newblock In {\em International Workshop on Internet and Network Economics},
  pages 477--489. Springer, 2008.

\bibitem{frbach1}
F.~Bach.
\newblock {Learning with Submodular functions: A convex Optimization
  Perspective (updated version)}.
\newblock {\em Arxiv}, 2013.

\bibitem{badanidiyuru2014streaming}
A.~Badanidiyuru, B.~Mirzasoleiman, A.~Karbasi, and A.~Krause.
\newblock Streaming submodular maximization: Massive data summarization on the
  fly.
\newblock In {\em Proceedings of the 20th ACM SIGKDD international conference
  on Knowledge discovery and data mining}, pages 671--680. ACM, 2014.

\bibitem{badanidiyuru2014fast}
A.~Badanidiyuru and J.~Vondr{\'a}k.
\newblock Fast algorithms for maximizing submodular functions.
\newblock {\em In SODA}, 2014.

\bibitem{balkanski2016learning}
E.~Balkanski, B.~Mirzasoleiman, A.~Krause, and Y.~Singer.
\newblock Learning sparse combinatorial representations via two-stage
  submodular maximization.
\newblock In {\em ICML}, pages 2207--2216, 2016.

\bibitem{bilmes2017deep}
J.~Bilmes and W.~Bai.
\newblock Deep submodular functions.
\newblock {\em arXiv preprint arXiv:1701.08939}, 2017.

\bibitem{feldman2012optimal}
N.~Buchbinder, M.~Feldman, J.~Naor, and R.~Schwartz.
\newblock A tight (1/2) linear-time approximation to unconstrained submodular
  maximization.
\newblock {\em In FOCS}, 2012.

\bibitem{chakrabarty2014provable}
D.~Chakrabarty, P.~Jain, and P.~Kothari.
\newblock Provable submodular minimization using wolfe's algorithm.
\newblock In {\em Advances in Neural Information Processing Systems}, pages
  802--809, 2014.

\bibitem{chakraborty2015adaptive}
S.~Chakraborty, O.~Tickoo, and R.~Iyer.
\newblock Adaptive keyframe selection for video summarization.
\newblock In {\em Applications of Computer Vision (WACV), 2015 IEEE Winter
  Conference on}, pages 702--709. IEEE, 2015.

\bibitem{das2012selecting}
A.~Das, A.~Dasgupta, and R.~Kumar.
\newblock Selecting diverse features via spectral regularization.
\newblock In {\em NIPS}, 2012.

\bibitem{dasgupta2013summarization}
A.~Dasgupta, R.~Kumar, and S.~Ravi.
\newblock Summarization through submodularity and dispersion.
\newblock In {\em ACL (1)}, pages 1014--1022, 2013.

\bibitem{el2009turning}
K.~El-Arini, G.~Veda, D.~Shahaf, and C.~Guestrin.
\newblock Turning down the noise in the blogosphere.
\newblock In {\em KDD}, 2009.

\bibitem{janvondrak}
U.~Feige, V.~Mirrokni, and J.~Vondr{\'a}k.
\newblock Maximizing non-monotone submodular functions.
\newblock {\em SIAM J. COMPUT.}, 40(4):1133--1155, 2007.

\bibitem{frank1956algorithm}
M.~Frank and P.~Wolfe.
\newblock An algorithm for quadratic programming.
\newblock {\em Naval research logistics quarterly}, 1956.

\bibitem{friedland2011submodular}
S.~Friedland and S.~Gaubert.
\newblock Submodular spectral functions of principal submatrices of a hermitian
  matrix, extensions and applications.
\newblock {\em Linear Algebra and its Applications}, 2011.

\bibitem{fujishige2005submodular}
S.~Fujishige.
\newblock {\em Submodular functions and optimization}, volume~58.
\newblock Elsevier Science, 2005.

\bibitem{fujishige2011submodular}
S.~Fujishige and S.~Isotani.
\newblock A submodular function minimization algorithm based on the
  minimum-norm base.
\newblock {\em Pacific Journal of Optimization}, 7:3--17, 2011.

\bibitem{goyal2011simpath}
A.~Goyal, W.~Lu, and L.~V. Lakshmanan.
\newblock Simpath: An efficient algorithm for influence maximization under the
  linear threshold model.
\newblock In {\em Data Mining (ICDM), 2011 IEEE 11th International Conference
  on}, pages 211--220. IEEE, 2011.

\bibitem{gu1994stable}
M.~Gu and S.~C. Eisenstat.
\newblock A stable and fast algorithm for updating the singular value
  decomposition.
\newblock {\em Yale University, New Haven, CT}, 1994.

\bibitem{hassani2017gradient}
H.~Hassani, M.~Soltanolkotabi, and A.~Karbasi.
\newblock Gradient methods for submodular maximization.
\newblock In {\em Advances in Neural Information Processing Systems}, pages
  5841--5851, 2017.

\bibitem{hassidim2016submodular}
A.~Hassidim and Y.~Singer.
\newblock Submodular optimization under noise.
\newblock {\em arXiv preprint arXiv:1601.03095}, 2016.

\bibitem{horel2016maximization}
T.~Horel and Y.~Singer.
\newblock Maximization of approximately submodular functions.
\newblock In {\em Advances in Neural Information Processing Systems}, pages
  3045--3053, 2016.

\bibitem{rkiyeruai2012}
R.~Iyer and J.~Bilmes.
\newblock Algorithms for approximate minimization of the difference between
  submodular functions, with applications.
\newblock {\em In UAI}, 2012.

\bibitem{rkiyersubmodBregman2012}
R.~Iyer and J.~Bilmes.
\newblock The submodular {B}regman and {L}ov\'asz-{B}regman divergences with
  applications.
\newblock In {\em NIPS}, 2012.

\bibitem{nipssubcons2013}
R.~Iyer and J.~Bilmes.
\newblock {Submodular Optimization with Submodular Cover and Submodular
  Knapsack Constraints}.
\newblock In {\em NIPS}, 2013.

\bibitem{curvaturemin}
R.~Iyer, S.~Jegelka, and J.~Bilmes.
\newblock {Curvature and Optimal Algorithms for Learning and Minimizing
  Submodular Functions }.
\newblock In {\em Neural Information Processing Society (NIPS)}, 2013.

\bibitem{rkiyersemiframework2013}
R.~Iyer, S.~Jegelka, and J.~Bilmes.
\newblock {Fast Semidifferential based Submodular function optimization}.
\newblock In {\em ICML}, 2013.

\bibitem{iyer2014monotone}
R.~Iyer, S.~Jegelka, and J.~Bilmes.
\newblock Monotone closure of relaxed constraints in submodular optimization:
  connections between minimization and maximization.
\newblock In {\em Proceedings of the Thirtieth Conference on Uncertainty in
  Artificial Intelligence}, pages 360--369. AUAI Press, 2014.

\bibitem{jegelka2011-nonsubmod-vision}
S.~Jegelka and J.~A. Bilmes.
\newblock Submodularity beyond submodular energies: coupling edges in graph
  cuts.
\newblock In {\em CVPR}, 2011.

\bibitem{karimi2017stochastic}
M.~Karimi, M.~Lucic, H.~Hassani, and A.~Krause.
\newblock Stochastic submodular maximization: The case of coverage functions.
\newblock In {\em Advances in Neural Information Processing Systems}, pages
  6853--6863, 2017.

\bibitem{kkt03}
D.~Kempe, J.~Kleinberg, and E.~Tardos.
\newblock Maximizing the spread of influence through a social network.
\newblock In {\em SIGKDD}, 2003.

\bibitem{krause2010sfo}
A.~Krause.
\newblock {SFO}: A toolbox for submodular function optimization.
\newblock {\em JMLR}, 11:1141--1144, 2010.

\bibitem{krause2008efficient}
A.~Krause, J.~Leskovec, C.~Guestrin, J.~VanBriesen, and C.~Faloutsos.
\newblock Efficient sensor placement optimization for securing large water
  distribution networks.
\newblock {\em Journal of Water Resources Planning and Management},
  134(6):516--526, 2008.

\bibitem{kulesza2012determinantal}
A.~Kulesza and B.~Taskar.
\newblock Determinantal point processes for machine learning.
\newblock {\em arXiv preprint arXiv:1207.6083}, 2012.

\bibitem{lin2012submodularity}
H.~Lin.
\newblock {\em Submodularity in Natural Language Processing: Algorithms and
  Applications}.
\newblock PhD thesis, University of Washington, Dept.\ of EE, 2012.

\bibitem{linbudget}
H.~Lin and J.~Bilmes.
\newblock Multi-document summarization via budgeted maximization of submodular
  functions.
\newblock {\em In NAACL}, 2010.

\bibitem{linacl}
H.~Lin and J.~Bilmes.
\newblock A class of submodular functions for document summarization.
\newblock {\em In ACL}, 2011.

\bibitem{lin2011optimal}
H.~Lin and J.~Bilmes.
\newblock Optimal selection of limited vocabulary speech corpora.
\newblock In {\em Interspeech}, 2011.

\bibitem{lin2009graph}
H.~Lin, J.~Bilmes, and S.~Xie.
\newblock Graph-based submodular selection for extractive summarization.
\newblock In {\em ASRU}, 2009.

\bibitem{liu2015svitchboard}
Y.~Liu, R.~Iyer, K.~Kirchhoff, and J.~Bilmes.
\newblock Svitchboard ii and fisver i: High-quality limited-complexity corpora
  of conversational english speech.
\newblock In {\em Sixteenth Annual Conference of the International Speech
  Communication Association}, 2015.

\bibitem{lovasz1983}
L.~Lov\'asz.
\newblock Submodular functions and convexity.
\newblock {\em Mathematical Programming}, 1983.

\bibitem{minoux1978accelerated}
M.~Minoux.
\newblock Accelerated greedy algorithms for maximizing submodular set
  functions.
\newblock {\em Optimization Techniques}, pages 234--243, 1978.

\bibitem{mirzasoleiman2015lazier}
B.~Mirzasoleiman, A.~Badanidiyuru, A.~Karbasi, J.~Vondr{\'a}k, and A.~Krause.
\newblock Lazier than lazy greedy.
\newblock In {\em AAAI}, pages 1812--1818, 2015.

\bibitem{mirzasoleiman13distributed}
B.~Mirzasoleiman, A.~Karbasi, R.~Sarkar, and A.~Krause.
\newblock Distributed submodular maximization: Identifying representative
  elements in massive data.
\newblock In {\em NIPS}, 2013.

\bibitem{mokhtari2017conditional}
A.~Mokhtari, H.~Hassani, and A.~Karbasi.
\newblock Conditional gradient method for stochastic submodular maximization:
  Closing the gap.
\newblock {\em arXiv preprint arXiv:1711.01660}, 2017.

\bibitem{narasimhanbilmes}
M.~Narasimhan and J.~Bilmes.
\newblock A submodular-supermodular procedure with applications to
  discriminative structure learning.
\newblock In {\em UAI}, 2005.

\bibitem{nemhauser1978}
G.~Nemhauser, L.~Wolsey, and M.~Fisher.
\newblock An analysis of approximations for maximizing submodular set
  functions---i.
\newblock {\em Mathematical Programming}, 14(1):265--294, 1978.

\bibitem{singla2016noisy}
A.~Singla, S.~Tschiatschek, and A.~Krause.
\newblock Noisy submodular maximization via adaptive sampling with applications
  to crowdsourced image collection summarization.
\newblock In {\em Thirtieth AAAI Conference on Artificial Intelligence}, 2016.

\bibitem{stan2017probabilistic}
S.~Stan, M.~Zadimoghaddam, A.~Krause, and A.~Karbasi.
\newblock Probabilistic submodular maximization in sub-linear time.
\newblock In {\em Proceedings of the 34th International Conference on Machine
  Learning-Volume 70}, pages 3241--3250. JMLR. org, 2017.

\bibitem{tschiatschek2014learning}
S.~Tschiatschek, R.~K. Iyer, H.~Wei, and J.~A. Bilmes.
\newblock Learning mixtures of submodular functions for image collection
  summarization.
\newblock In {\em Advances in neural information processing systems}, pages
  1413--1421, 2014.

\bibitem{wei2014fast}
K.~Wei, R.~Iyer, and J.~Bilmes.
\newblock Fast multi-stage submodular maximization.
\newblock In {\em ICML}, 2014.

\bibitem{wei2015submodularity}
K.~Wei, R.~Iyer, and J.~Bilmes.
\newblock Submodularity in data subset selection and active learning.
\newblock In {\em International Conference on Machine Learning}, pages
  1954--1963, 2015.

\bibitem{wei2014submodular}
K.~Wei, Y.~Liu, K.~Kirchhoff, C.~Bartels, and J.~Bilmes.
\newblock Submodular subset selection for large-scale speech training data.
\newblock {\em Proceedings of ICASSP, Florence, Italy}, 2014.

\bibitem{wei2013using}
K.~Wei, Y.~Liu, K.~Kirchhoff, and J.~Bilmes.
\newblock Using document summarization techniques for speech data subset
  selection.
\newblock In {\em NAACL-HLT}, 2013.

\bibitem{wolsey1982analysis}
L.~A. Wolsey.
\newblock An analysis of the greedy algorithm for the submodular set covering
  problem.
\newblock {\em Combinatorica}, 2(4):385--393, 1982.

\end{thebibliography}
\end{document}